\documentclass[a4paper,fleqn]{cas-dc}

\usepackage[authoryear]{natbib}
\usepackage{subfig}
\usepackage{enumitem}
\usepackage{caption}
\usepackage{hyperref}
\usepackage{cleveref}
\usepackage{amssymb}
\usepackage{amsmath}
\usepackage{xcolor}
\usepackage{bm}
\usepackage{array}
\usepackage{pifont}
\crefname{table}{Table}{Tables}
\crefformat{table}{#2Table\,#1#3} 

\crefname{figure}{}{}  
\Crefname{figure}{}{} 
\creflabelformat{figure}{#2Fig.~#1#3} 
\crefname{section}{Section}{Sections}  

\crefname{equation}{}{}  
\Crefname{equation}{}{} 
\creflabelformat{equation}{#2Eq.~#1#3} 

\def\tsc#1{\csdef{#1}{\textsc{\lowercase{#1}}\xspace}}
\tsc{WGM}
\tsc{QE}

\usepackage{fancyhdr}
\begin{document}
\let\WriteBookmarks\relax
\def\floatpagepagefraction{1}
\def\textpagefraction{.001}

\shorttitle{}    

\shortauthors{}  

\title [mode = title]{A Fully Automatic Framework for Intracranial Pressure Grading: Integrating Keyframe Identification, ONSD Measurement and Clinical Data}  

\tnotemark[1] 

\tnotetext[1]{*Co-first authors. **Corresponding authors.} 

%

\author[1]{{Pengxu Wen}}[orcid=0009-0000-5211-4876]
\fnmark[*]
\ead{pengxuwen@smail.nju.edu.cn}
\credit{Writing – review \& editing, Writing -- original draft, Visualization, Validation, Software, Methodology, Data curation, Conceptualization}

\author[2]{{Tingting Yu}}
\fnmark[*]
\ead{cathytina_19820515@163.com}
\credit{Writing -- original draft, Validation, Methodology, Resources, Data curation, Funding acquisition}

\author[1]{{Ziwei Nie}}[orcid=0000-0002-4499-7207]
\fnmark[**]
\ead{nieziwei@nju.edu.cn}
\credit{Writing -- review \& editing, Methodology, Supervision, Conceptualization, Funding acquisition}

\author[2]{{Cheng Jiang}}
\fnmark[**]
\ead{15951756001@163.com}
\credit{Validation, Resources, Data curation, Methodology, Supervision}

\author[2]{{Zhenyu Yin}}
\credit{Validation, Resources, Data curation}

\author[3]{{Mingyang He}}
\credit{Validation, Resources, Data curation}

\author[4]{{Bo Liao}}
\credit{Validation, Resources, Data curation}

\author[1]{{Xiaoping Yang}}[orcid=0000-0002-8298-1273]
\fnmark[**]
\ead{xpyang@nju.edu.cn}
\credit{Writing – review \& editing, Methodology, Supervision, Funding acquisition}

\address[a]{School of Mathematics, Nanjing University, Nanjing, 210093, China}
\address[b]{Department of Geriatric, Nanjing Drum Tower Hospital Affiliated of Nanjing University Medical School, Nanjing, 210008, China}
\address[c]{Drum Tower Clinical Medical College, Nanjing University of Chinese Medicine, Nanjing, 210023, China}
\address[d]{Department of Intensive Medicine, The Affiliated Jiangning Hospital with Nanjing Medical University, Nanjing, 211100, China}



\begin{abstract}
Intracranial pressure (ICP) elevation poses severe threats to cerebral function, thus necessitating monitoring for timely intervention. While lumbar puncture is the gold standard for ICP measurement, its invasiveness and associated risks drive the need for non-invasive alternatives. Optic nerve sheath diameter (ONSD) has emerged as a promising biomarker, as elevated ICP directly correlates with increased ONSD. However, current clinical practices for ONSD measurement suffer from inconsistency in manual operation, subjectivity in optimal view selection, and variability in thresholding, limiting their reliability. To address these challenges, we introduce a fully automatic two-stage framework for ICP grading, integrating keyframe identification, ONSD measurement and clinical data. Specifically, the fundus ultrasound video processing stage performs frame-level anatomical segmentation, rule-based keyframe identification guided by an international consensus statement, and precise ONSD measurement. The intracranial pressure grading stage then fuses ONSD metrics with clinical features to enable the prediction of ICP grades, thereby demonstrating an innovative blend of interpretable ultrasound analysis and multi-source data integration for objective clinical evaluation. Experimental results demonstrate that our method achieves a validation accuracy of $0.845 \pm 0.071$ (with standard deviation from five-fold cross-validation) and an independent test accuracy of 0.786, significantly outperforming conventional threshold-based method ($0.637 \pm 0.111$ validation accuracy, $0.429$ test accuracy).  Through effectively reducing operator variability and integrating multi-source information, our framework establishes a reliable non-invasive approach for clinical ICP evaluation, holding promise for improving patient management in acute neurological conditions.
\nocite{*}
\end{abstract}

\begin{graphicalabstract}
\includegraphics[width=\textwidth]{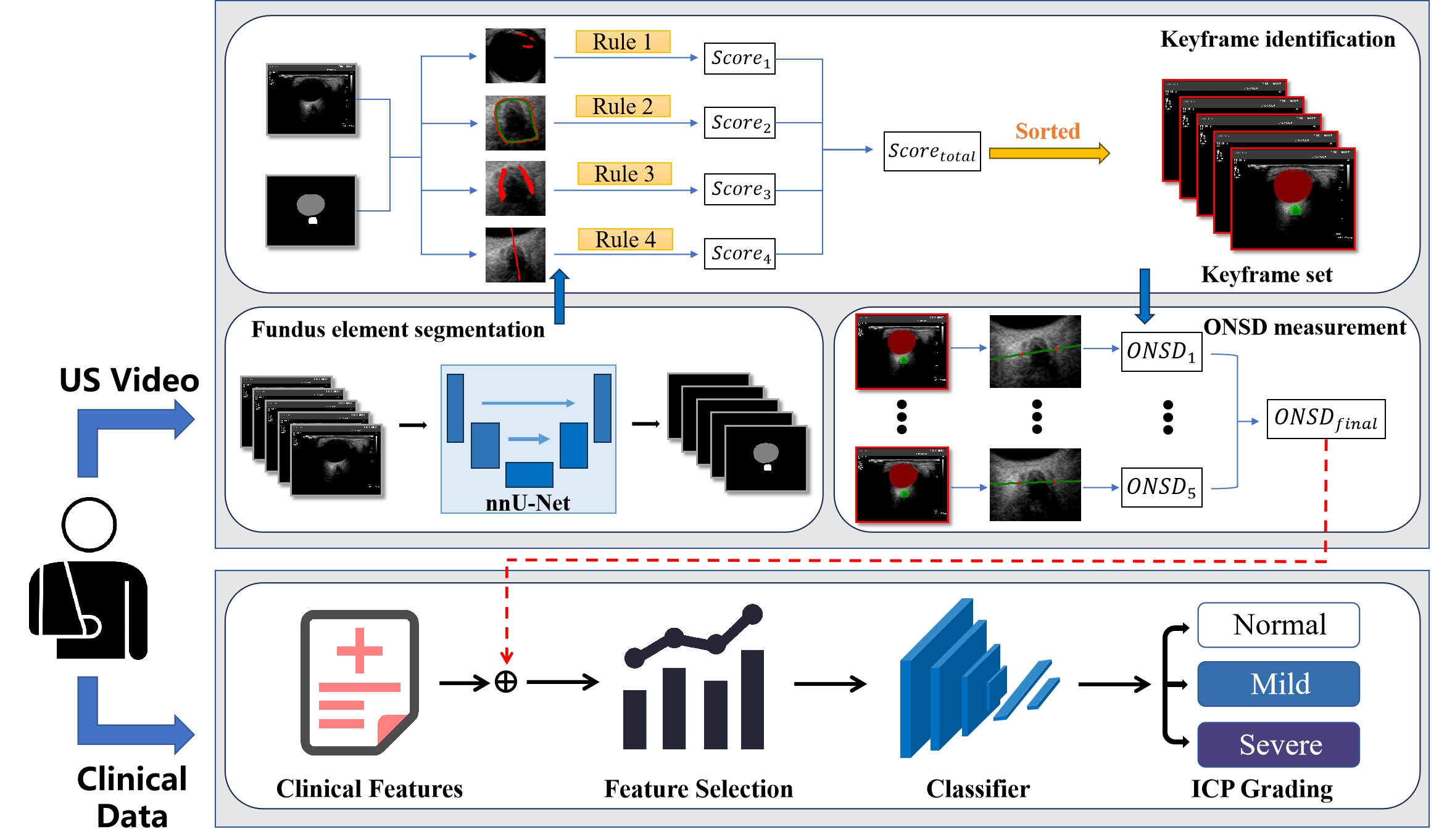}
\end{graphicalabstract}

\begin{highlights}
\item We develop a fully automatic framework that can select keyframes from ocular fundus ultrasound videos, precisely measure optic nerve sheath diameter (ONSD) on the selected frames, and integrate clinical data to provide accurate grading of  intracranial pressure (ICP).
\item To our knowledge, we are the first to develop a quantitative interpretable algorithm for keyframe identification in ocular fundus ultrasound videos, which is designed in strict accordance with an international consensus statement on optic nerve sheath diameter imaging and measurement.
\item To our knowledge, this is the first work to combine ONSD with patients' clinical information for ICP grading.
\item Our model achieves an ICP grading accuracy of $0.845 \pm 0.071$ on the validation set and $0.786$ on an independent test set, demonstrating significant superiority over conventional threshold-based method ($0.637 \pm 0.111$ validation accuracy, $0.429$ test accuracy).
\end{highlights}


\begin{keywords}
 Fundus ultrasound \\ 
 ICP grading  \\
 Keyframe identification \\
 ONSD measurement \\
 Clinical Data Fusion
\end{keywords}

\maketitle

\section{Introduction}
\label{intro}
Elevated intracranial pressure (ICP) has severe detrimental effects on cerebral function, potentially restricting blood flow to the brain, leading to oxygen deficiency and nutrient insufficiency, thereby causing irreversible brain damage.
Investigation and continuous monitoring of ICP elevation are critical for enabling timely clinical interventions and improving patient prognosis \citep{Hawryluk_2022}.
In clinical practice, lumbar puncture serves as a commonly employed invasive procedure for measuring cerebrospinal fluid pressure to accurately assess ICP \citep{Evensen_2020}. However, during this procedure, patients may suffer from significant discomfort or pain, while requiring monitoring for potential risks including infection, hemorrhage, and post-puncture headache.
Ocular fundus ultrasonographic imaging provides a safer and more convenient alternative as a non-invasive ICP measurement technique \citep{Moretti_2009,Robba_2017}. As the optic nerve sheath constitutes a continuation of the intracranial dura mater and subarachnoid space, elevated intracranial pressure directly induces an increase in optic nerve sheath diameter, thus the optic nerve sheath diameter (ONSD) is recognized as a significant parameter for predicting elevated ICP \citep{Liu_1993,Killer_2007,Moore_2023}.

During ocular fundus ultrasound examinations, sonographers are required to manually operate the computer for optimal view selection while maintaining one-handed control of the ultrasound probe. Subsequently, on the selected optimal view, manual measurement markers are placed 3 mm posterior to the eyeball along the axis perpendicular to the optic nerve to determine the ONSD \citep{Hirzallah_2024}. The measured value is then compared with the standard ONSD threshold, which when exceeded indicates significant elevation of intracranial pressure.
The limitations of current clinical technique include operator-dependent variability, subjectivity in optimal view selection, measurement instability of the optic nerve sheath diameter, and inconsistency in standard thresholds\citep{Stevens_2021_optic}.

Given the presence of numerous objective and subjective confounding factors in clinical examinations, current workflows remain depending on experienced physicians. Therefore, the purpose of this study is to develop a fully automatic framework capable of autonomously selecting keyframes from ocular fundus ultrasound videos, precisely measuring ONSD, and ultimately conduct graded prediction of ICP through integration of patients' clinical information. The main contributions of our work can be summarized as follows:
\begin{itemize}
    \item 
    We develop a fully automatic framework that can select keyframes from ocular fundus ultrasound videos, precisely measure ONSD on the selected frames, and integrate clinical data to provide accurate grading of ICP.
    \item To our knowledge, we are the first to develop a quantitative interpretable algorithm for keyframe identification in ocular fundus ultrasound videos, which is designed in strict accordance with an international consensus statement on optic nerve sheath diameter imaging and measurement. 
    \item To our knowledge, this is the first work to combine ONSD with patients' clinical information for ICP grading.
    \item Our model achieves an ICP grading accuracy of $0.845 \pm 0.071$ on the validation set and $0.786$ on an independent test set, demonstrating significant superiority over conventional threshold-based method ($0.637 \pm 0.111$ validation accuracy, $0.429$ test accuracy).
\end{itemize}

\section{Related works}

\subsection{Segmentation of optic nerve sheath in ultrasound images}

To precisely measure the optic nerve sheath diameter, numerous studies have been dedicated to segmenting ultrasound images using automatic or semi-automatic segmentation methods \citep{Gerber_2017,Soroushmehr_2019,Meiburger_2020,Stevens_2021_an,Rajajee_2021}. However, all these methods rely on conventional image processing techniques,  which are typically fine-tuned for specific databases exclusively used as training and testing sets \citep{Meiburger_2021}. Deep learning models have recently seen extensive applications in medical image segmentation tasks. Among them, the U-Net architecture has emerged as a widely adopted solution for such tasks \citep{Ronneberger_2015}. In 2021, \citeauthor{Meiburger_2021} (\citeyear{Meiburger_2021}) proposed a U-Net-based automatic segmentation method for the optic nerve sheath in ultrasound images. Through this method, they demonstrated that deep learning (DL) approaches exhibit clear advantages over previous techniques, achieving a mean absolute error of 0.48 mm in ONSD measurements. In 2023,  \citeauthor{Marzola_2023} (\citeyear{Marzola_2023}) also trained a U-NET to segment the optic nerve sheath for ONSD measurement, achieving a Dice score of $0.719 \pm 0.139$ via five-fold cross-validation on a dataset of 464 manually selected ultrasound images.

\subsection{Keyframe identification for ultrasound videos}\label{Sec. 2.2}

For precise measurement of the optic nerve sheath diameter, it is imperative to select the most representative views from ultrasound video sequences (i.e. key frames) to ensure measurement accuracy. Contemporary deep neural network models for video key frame detection can generally be categorized into two principal methodologies. The first approach formulates key frame detection as a binary classification problem, where detection models are trained using cross-entropy loss functions to classify each video frame as either a key frame or non-key frame \citep{Ciusdel_2020}. This methodology leverages the advantages of classification algorithms to extract image features from individual frames and generate categorical predictions. The second approach treats key frame detection as a regression problem, estimating a relevance or importance score for each frame to indicate its likelihood of selection as a key frame (\citeauthor{WangA_2021}, \citeyear{WangA_2021}; \citeauthor{WangB_2022}, \citeyear{WangB_2022}). Such models are trained with mean squared error loss functions, enabling refined evaluation of frame-level significance. Regression-based methods demonstrate enhanced flexibility in representing relative importance among video frames\citep{Feng_2024}.

However, both categories of deep learning models necessitate labeling of all frames in video sequences during training. This requirement is especially stringent for regression-based deep neural networks, as they demand precise quantitative scoring of every individual frame. Such a process imposes substantial clinical challenges and time burdens on physicians. \citeauthor{Huang_2022} (\citeyear{Huang_2022}) proposed a reinforcement learning-based framework to extract keyframes from breast ultrasound videos by scoring eight image metrics by integrating anatomical and diagnostic lesion characteristics into the keyframe search process, providing a new methodological direction for keyframe detection tasks. However, The reinforcement learning approach relies on high-quality annotations, which makes it difficult to achieve high efficiency in video scenarios. \citeauthor{Singh_2022} (\citeyear{Singh_2022}) employed an object detection network to identify keyframes containing the optic nerve sheath in ultrasound videos and subsequently measured ONSD. However, the keyframe definition criteria in Singh et al.'s study may be considered oversimplified, since frames displaying clear optic nerve sheath are categorized as optimal.

\begin{figure*}[htbp]
  \centering
   \includegraphics[width=0.9\textwidth]{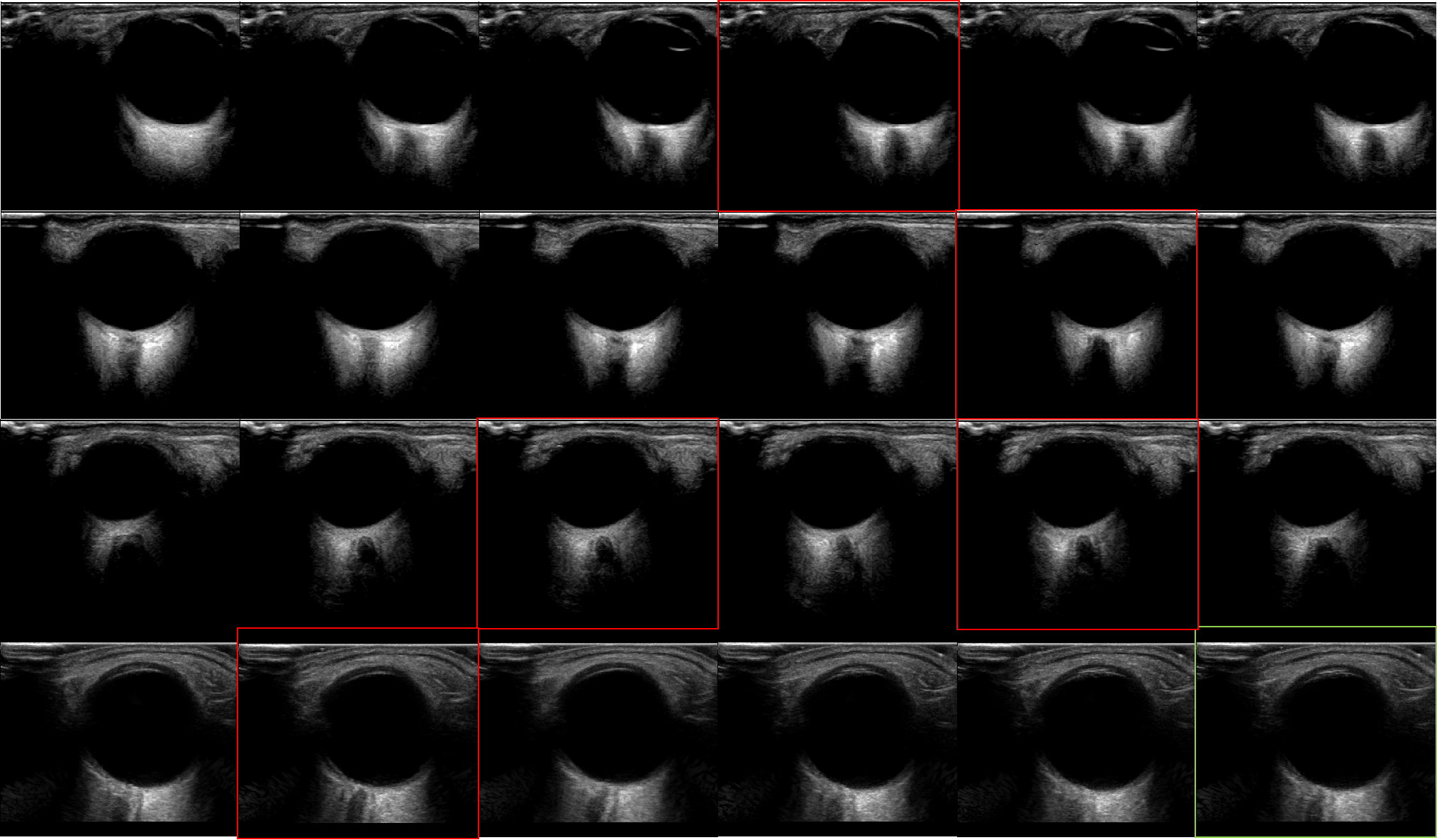}
    \caption{Examples of fundus ultrasound videos. The keyframes of different patients appear at distinct time points in the videos, highlighted in red (each row represents the fundus ultrasound video of one patient). The keyframe of each video is defined relative to its own content, and keyframes across different videos are not comparable (e.g., the red-highlighted keyframe in the first row exhibit notably lower image quality than the green-marked suboptimal frame in the fourth row), posing significant challenges for keyframe identification.}\label{fig1}
\end{figure*}

In summary, this study confronts the following technical challenges in keyframe identification for fundus ultrasound videos: Firstly, due to the individual heterogeneity of the optic nerve anatomical structure and subjective differences in physician operations, the keyframe identification has intra case relativity and inter case incomparability, as shown in \cref{fig1}, which makes it difficult to simply treat this task as a binary classification problem. Secondly, regression-based methods require large-scale annotated data for training, yet clinicians struggle to objectively and quantitatively score video frames. Finally, existing deep learning methods lack interpretability in key frame selection decisions, failing to support clinical decision credibility verification. 

\subsection{Measurement of ONSD and elevation of ICP}
After selecting the most representative frames from ultrasound videos, precise measurement of ONSD in keyframes becomes critical. This process imposes stringent demands on clinicians' experience and refined operational techniques. However, ocular ultrasound images frequently exhibit suboptimal quality characterized by low signal-to-noise ratio, insufficient contrast, and ill-defined ONS boundaries, all of which compromise measurement accuracy. Furthermore, when measuring ONSD 3 mm below the eyeball, clinicians are required to manually trace 3 mm along the ONS towards the globe, then delineate a perpendicular line to calculate the intersected distance between ONS boundaries. This procedure typically utilizes electronic calipers available on the screen of an ultrasound machine. The manually measured ONSD value is not accurate due to the orientation limitation of the electronic caliper \citep{Singh_2022}. Moreover, due to numerous operational interferences and methodological limitations in manual ONSD measurement, significant inter-observer and intra-observer variability persists in measurement reproducibility and stability. To address these challenges, some studies have proposed automatic approaches for ONSD measurement. \citeauthor{Soroushmehr_2019}(\citeyear{Soroushmehr_2019}) utilized super-pixel analysis with guided filtering and linear integral analysis, automating ONSD measurement by analyzing intensity derivatives at the 3-mm position behind the eyeball. \citeauthor{Rajajee_2021} (\citeyear{Rajajee_2021})  developed an algorithm integrating clustering and image segmentation to quantify ONSD from ultrasound images. \citeauthor{Singh_2022} (\citeyear{Singh_2022})  measured ONSD by drawing parabolas from the optic nerve tip, selecting the best-fitted one via intensity summation variation, and using it to quantify ONSD at 3 mm along the optic nerve axis. \citeauthor{li_2025} (\citeyear{li_2025}) proposed a framework employing Gaussian Mixture Model (GMM)-based localization and KL-divergence boundary refinement for ONSD measurement. 

In addition, although numerous studies have demonstrated significant correlations between ONSD and elevated ICP \citep{Del_2016, Jeub_2020, Berhanu_2023}, diagnostic thresholds for identifying intracranial hypertension exhibit marked heterogeneity across studies (ranging from 4.1 to 7.2 mm) \citep{Berhanu_2023, Stevens_2021_optic}, primarily attributable to variations in equipment specifications and operator expertise. Current clinical practice relies solely on single ONSD thresholds for binary classification of ICP (i.e., values above the threshold are designated as intracranial hypertension) without incorporating other patient-specific clinical features. To enhance clinical applicability, this study aims to develop a prediction model for three-tiered grading of ICP (i.e., normal, mild intracranial hypertension, and severe intracranial hypertension) by integrating ONSD measurements from ocular ultrasound videos with multidimensional patient clinical profiles.

\section{Proposed Methods}

\subsection{The overall framework}
\begin{figure*}[htbp]
  \centering
   \includegraphics[width=0.9\textwidth]{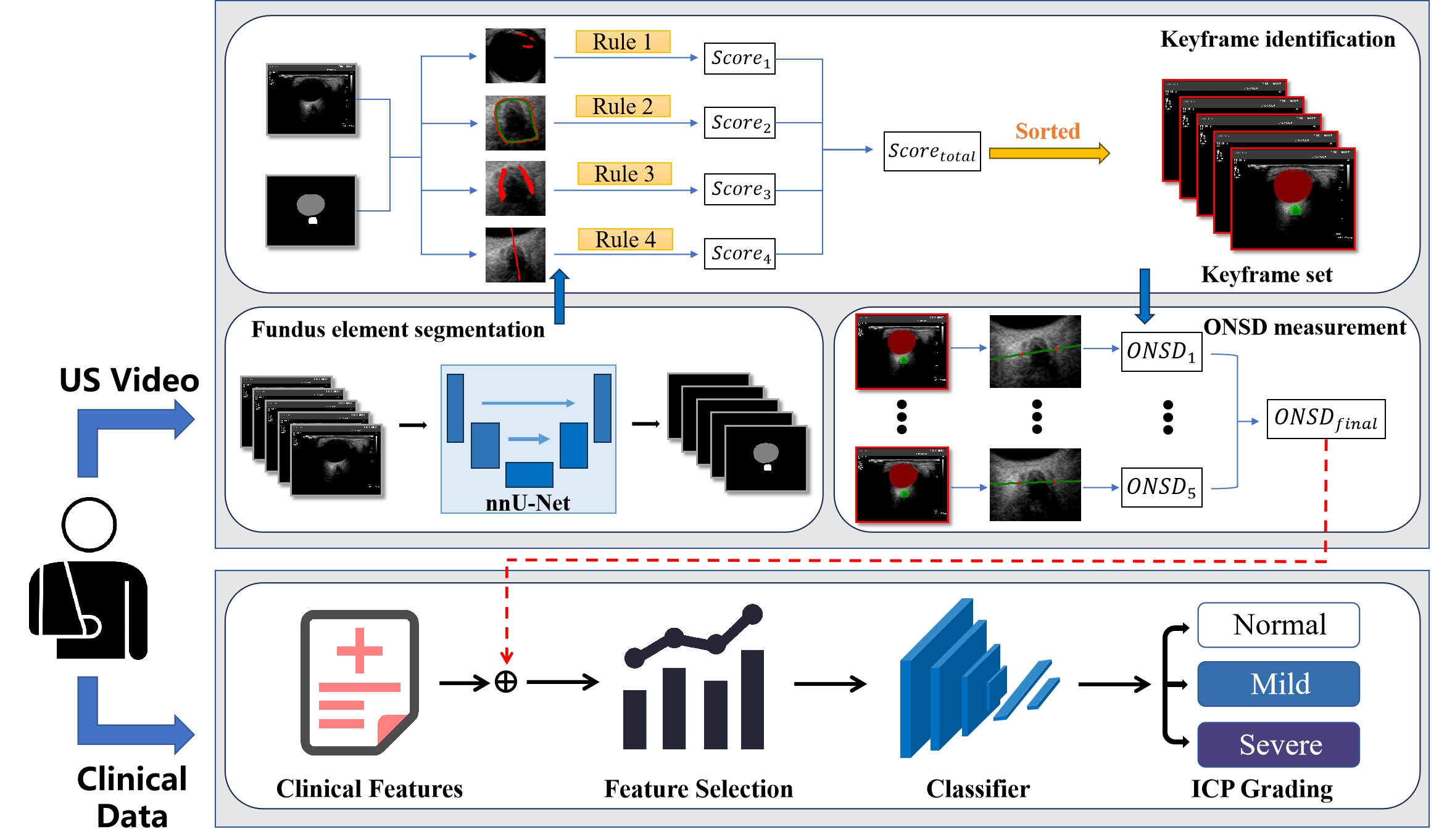}
    \caption{Pipeline of the proposed framework. The framework comprises a fundus ultrasound video processing stage and an ICP grading stage. In the fundus ultrasound video processing stage, the patient’s ultrasound video is input to automatically identify keyframes and measure the ONSD. The ICP grading stage uses the patient’s clinical information and the ONSD obtained from the previous stage as inputs to grade the patient’s ICP. }\label{fig2}
\end{figure*}

As illustrated in \cref{fig2}, this study presents a fully automatic framework for measuring the optimal-frame ONSD from ultrasound videos and performing hierarchical prediction of ICP by integrating patients' clinical information. The framework is fundamentally divided into two stages: the fundus ultrasound video processing stage and the intracranial pressure grading stage.

In the fundus ultrasound video processing stage, the workflow is as follows: First, the patient’s fundus ultrasound video is input into the fundus element segmentation module, where each frame of the video undergoes elemental segmentation to identify the eyeball and optic nerve sheath. Second, to enable the interpretable identification of keyframes, this study independently summarizes four qualitative rules for selecting keyframes based on the technical specifications proposed in \textit{An International Consensus Statement on Optic Nerve Sheath Diameter Imaging and Measurement} \citep{Hirzallah_2024}. Through quantitative analysis, a quantifiable and interpretable scoring algorithm is designed in the keyframe identification module, which assigns interpretable scores to each frame based on the ultrasound images and their corresponding segmentation masks. The top five highest-scoring frames are then selected as the keyframe set. Finally, in the ONSD measurement module, the ONSD values at 3 mm below the eyeball for each frame in the keyframe set are measured according to the designed strategy, and abnormal outliers are removed to obtain the precise ONSD value for the entire ultrasound video.

In the intracranial pressure grading stage, the process commences with extracting clinical features which potentially influence ICP from the patient’s medical records. The ONSD feature obtained from the fundus ultrasound video processing stage is fused with these clinical features; subsequently, feature screening is performed to eliminate variables irrelevant to ICP. Ultimately, the filtered features are fed into a classifier to complete the hierarchical prediction of intracranial pressure.  

The details of each stage are illustrated below.

\subsection{Fundus ultrasound videos processing stage}

\subsubsection{Fundus element segmentation module}

In the fundus element segmentation module, this study employs the advanced medical image segmentation framework nnU-Net \citep{Isensee_2021} to segment the eyeball and optic nerve sheath regions in each frame of the video. To prevent data leakage arising from the high similarity between frames within the same video, we ensure during dataset preparation that frames from each patient are exclusively allocated to either the training set or the test set. During the training of the 2D nnU-Net, the Dice loss and cross-entropy loss are used as joint loss functions, with the equilibrium coefficient set to 1:1. 

\subsubsection{Keyframe identification module}

In the keyframe identification module, this study designs an interpretable scoring algorithm based on the aforementioned international consensus statement to screen five keyframes from each video as the keyframe set. Specifically, four qualitative keyframe selection rules are first summarized from the consensus: "Presence or absence of the lens and anterior chamber", "Edge clarity of the optic nerve sheath", "Salience of the 'two-bright-three-dark' structure", and "Degree of vertical display of the optic nerve sheath". Corresponding features are then extracted in combination with the segmentation results to quantify these rules. The first rule, "Presence or absence of the lens and anterior chamber", serves as a prerequisite for keyframe screening. Frames containing lens or anterior chamber structures are preferentially excluded from consideration. The second rule, "Edge clarity of the optic nerve sheath", uses the edge clarity of the optic nerve sheath region as the evaluation indicator, and frames with clearer edges have higher priority. In the third rule, "Integrity of the 'two-bright-three-dark' structure", "two-bright-three-dark" refers to the alternating bright and dark regions in the optic nerve sheath area, where the "two-bright" parts correspond to the subarachnoid region and the "three-dark" parts correspond to the dura mater and optic nerve regions. Frames with more pronounced light-dark contrast are more in line with the selection criteria. The fourth rule, "Degree of vertical display of the optic nerve sheath", assesses the overall orientation of the optic nerve sheath. Frames in which the optic nerve sheath appears more vertically oriented are prioritized.

\begin{figure}[htbp]
\centering
\subfloat[]{
  \includegraphics[width=0.21\columnwidth]{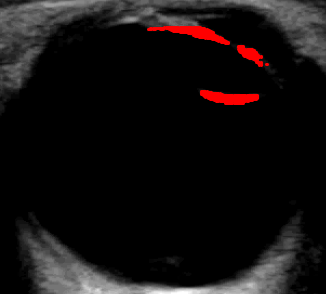}
  \label{fig3a}
}
\hfill
\subfloat[]{
  \includegraphics[width=0.21\columnwidth]{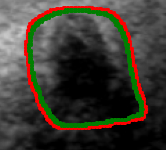}
  \label{fig3b}
}
\hfill
\subfloat[]{
  \includegraphics[width=0.21\columnwidth]{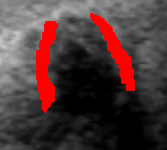}
  \label{fig3c}
}
\hfill
\subfloat[]{
  \includegraphics[width=0.20\columnwidth]{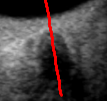}
  \label{fig3d}
}

\caption{Visualization of the quantification of four qualitative rules. (a) Rule 1: "Presence or absence of the lens and anterior chamber"—the red area indicates the lens and anterior chamber structures. (b) Rule 2: "Edge clarity of the optic nerve sheath"—the red area represents the external region of the optic nerve sheath edge, and the green area represents the internal region of the optic nerve sheath edge. (c) Rule 3: "Salience of the 'two-bright-three-dark' structure"—the red area corresponds to the "two-bright" structures (subarachnoid space). (d) Rule 4: "Degree of vertical display of the optic nerve sheath"—the red line denotes the central line of the optic nerve sheath.}\label{fig3}
\end{figure}

Building on these four rules, the following detailed quantification strategies are developed for each criterion, structured as itemized steps with corresponding technical descriptions:

\begin{itemize}
\item{\textbf{Presence or absence of the lens and anterior chamber (Rule 1)}:  
  Since the lens and anterior chamber are located inside the eyeball, frames containing these structures exhibit high-intensity regions within the intra-ocular area (as shown in \cref{fig3a}). The quantification strategy is as follows:  
  \begin{itemize}
    \item Map the eyeball segmentation mask to the original image to obtain the eyeball region.  
    \item Erode the region inward by 1.5 mm using an erosion operator to exclude contour interference.  
    \item Calculate the sum of all pixel values in the eroded image as the quantitative indicator $\operatorname{score}_{1}$.  
  \end{itemize}
  Larger $\operatorname{score}_{1}$ indicates more prominent lens/anterior chamber structures and lower priority.}

\item{\textbf{Edge clarity of the optic nerve sheath (Rule 2)}:  
  This rule evaluates the grayscale changes in the edge region of the optic nerve sheath (as shown in \cref{fig3b}). The quantification strategy is as follows:  
  \begin{itemize}
    \item Erode the optic nerve sheath segmentation mask inward and dilate it outward by 0.1 mm, respectively.  
    \item Subtract the original mask to obtain segmentation masks covering a band region of 0.2 mm width, with 0.1 mm inside and 0.1 mm outside the contour.  
    \item Map these masks to the original image and calculate the difference in average grayscale values as $\operatorname{score}_{2}$.  
  \end{itemize}
  Larger $\operatorname{score}_{2}$ signifies more drastic grayscale changes at the sheath edges and higher priority.}

\item{\textbf{Saliency of the 'two-bright-three-dark' structure (Rule 3)}:  
  The "two-bright-three-dark" structure (subarachnoid bright regions and dura mater--optic nerve dark regions) aligns with the sheath's overall trend and is prominent 3 to 5 mm below the eyeball (as shown in \cref{fig3c}). The quantification strategy is as follows:  
  \begin{itemize}
    \item Map eyeball and sheath masks to extract regions of interest.  
    \item Calculate the sheath central line, identify its proximal intersection with the eyeball, and extract a 3 to 5 mm segment below this intersection point.  
    \item Sum grayscale projections perpendicular to the central line to generate a feature vector.  
    \item Compute the sum of grayscale differences between peaks and valleys in this vector as $\operatorname{score}_{3}$.  
  \end{itemize}
  Larger $\operatorname{score}_{3}$ indicates stronger structural salience and higher priority.}

\item{\textbf{Degree of vertical display of the optic nerve sheath (Rule 4)}:  
  The overall trend of the optic nerve sheath is characterized by its central line (as shown in \cref{fig3d}). The quantification strategy is as follows:  
  \begin{itemize}
    \item Map the optic nerve sheath segmentation mask to the original image to obtain the sheath region.  
    \item Calculate the central line and measure its angle $\theta$ with the vertical direction.  
    \item Use $\tan\theta$ as the quantitative indicator $\operatorname{score}_{4}$.  
  \end{itemize}
  Larger $\operatorname{score}_{4}$ indicates a more horizontal orientation and lower priority.}
\end{itemize}

In response to the technical challenges outlined in \cref{Sec. 2.2} regarding the identification of keyframes in fundus ultrasound videos, this study introduces an interpretable scoring algorithm to evaluate each frame based on the aforementioned four quantitative metrics. The scoring model is formulated as follows:
\begin{equation}
\operatorname{score}_{\text {total }}=w_{1} \operatorname{score}_{1}+w_{2} \operatorname{score}_{2}+w_{3} \operatorname{score}_{3}+w_{4} \operatorname{score}_{4}
\label{Eq.1}
\end{equation}
where each metric is assigned a distinct weight coefficient, and the final score is computed through a weighted average. To calibrate these coefficients, supervised signals from expert ophthalmologists are incorporated. Despite the intra-patient correlation and inter-case heterogeneity inherent in keyframe identification, clinicians can reliably identify representative high-quality and suboptimal frames within individual video sequences. Leveraging such labeled examples, a Linear Discriminant Analysis (LDA) \citep{Pedregosa_2011} is performed to determine the optimal weights: \(w_1 = -0.2873\), \(w_2 = 0.3585\), \(w_3 = 0.1115\), and \(w_4 = -0.1179\). These coefficients are then applied to the respective metrics to generate the final keyframe identification score $\operatorname{score}_{\text {total }}$ for each frame. Frames are subsequently ranked in descending order of their scores, and the top five frames are selected as the keyframe set for the corresponding video.


\subsubsection{ONSD measurement module}

\begin{figure}[htbp]
  \centering
   \includegraphics[width=0.8\columnwidth]{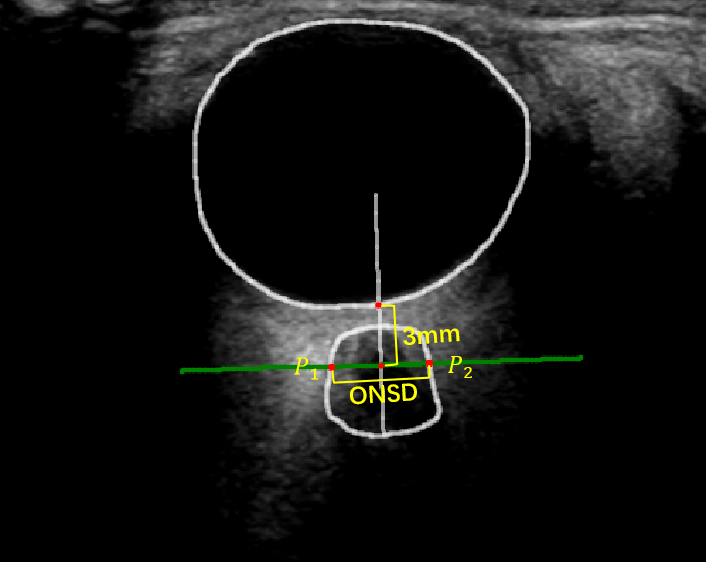}
    \caption{Schematic illustration of the ONSD measurement method.}\label{fig4}
\end{figure}

In the ONSD measurement module, this study develops a standardized measurement pipeline to compute the optic nerve sheath diameter from the keyframe set, integrating geometric analysis and statistical outlier handling. Specifically, the workflow for measuring the optic nerve sheath diameter in each image of the keyframe set is as follows: First, leveraging the segmentation results from the prior module, the central line of the optic nerve sheath is extracted based on its binary mask, and the proximal intersection point of this central line with the eyeball region is precisely determined. Subsequently, using the image's pixel spacing information, a target point is located 3 mm downward along the central line from the intersection point. A perpendicular line to the central line is then constructed through this target point to obtain the intersection points (\(P_1\) and \(P_2\)) with the bilateral edges of the optic nerve sheath. Finally, the Euclidean distance between \(P_1\) and \(P_2\) is calculated to derive the optic nerve sheath diameter for the single frame(as shown in \cref{fig4}).

For the five ONSD measurements corresponding to the keyframe set of each video, this study employs the Interquartile Range (IQR) method for outlier detection and removal. Specifically, the first quartile (Q1), third quartile (Q3), and interquartile range (IQR = Q3 - Q1) of the data are calculated to define a reasonable range of [Q1 - 1.5 \ding{53} IQR, Q3 + 1.5 \ding{53} IQR]. Data points outside this range are identified as outliers and excluded. After outlier processing, the maximum value among the remaining data is selected as the final ONSD measurement for the video.

\subsection{ICP grading stage}

In the ICP grading stage, this study develops a multi-dimensional precise ICP grading model integrating patient clinical characteristics and ONSD. Specifically, according to international standards \citep{My_2016}, patient ICP  (unit: \(\mathrm{mm\ H_2O}\)) is categorized into three grades: 80 to 180 as normal ICP, 180 to 280 as mild intracranial hypertension, and >280 as severe intracranial hypertension (indicating risk of brain herniation). The overall workflow of this stage is as follows: First, 49 potential clinical characteristics associated with ICP (including gender, age, blood routine indices, liver and renal function, etc.) are extracted and selected from patient clinical records, which are combined with the bilateral ocular mean ONSD features obtained from the fundus ultrasound video processing stage to construct a 50-dimensional comprehensive feature vector. Subsequently, Lasso regression is applied for feature selection on these 50 features, retaining 14 features with significant predictive value. Finally, the selected features are input into eight machine learning classifiers—Logistic regression, Decision Tree, Random Forest, SVM, KNN, XGBoost, MLP, and Bayes—for five-fold cross-validation prediction of ICP grades.

\section{Experiments}
\begin{table*}[htbp]
\caption{Summary of dataset information for different modules. Details include the number of patients, data modality, and the split of training and test data.} \label{tal0}
\centering
\begin{tabular*}{\textwidth}{@{\extracolsep{\fill}}lccccc@{}}
\toprule
Module & Patients & Modality & Training Data & Test Data \\
\midrule
Fundus Element Segmentation  & $64$ & Image & $609$ frames & $153$ frames \\
Keyframe Identification & $64$ & Image & $40$ videos & $24$ videos \\
ICP Grading & $51$ & Image \& Clinical Information & $102$ videos & $28$ videos \\
\bottomrule
\end{tabular*}
\end{table*}

\subsection{dataset}
In this study, patients who underwent lumbar puncture to obtain real ICP values are included, while those with ultrasound videos showing severe artifacts or poor imaging quality are excluded. Based on these criteria, a total of 208 ocular fundus ultrasound videos from 64 patients (24 males and 40 females, mean age 57.33±8.91 years) are retrospectively collected from Nanjing Drum Tower Hospital between January 2021 and May 2025. Furthermore, an independent test set consisting of 38 videos from 18 patients is prospectively collected at the same institution from May to June 2025, incorporating both video and clinical data.  Images stored in DICOM format are acquired by a senior physician using two different scanner models. The original dataset features frame counts per video ranging from 22 to 640 frames, with an average of 200 frames, and two resolution formats (1024 × 768 pixels with spacing [0.065, 0.065] and 800 × 600 pixels with spacing [0.083, 0.083]), while the independent test set includes two resolution configurations (1182 × 899 pixels with spacing [0.056, 0.056] and 1260 × 910 pixels with spacing [0.067, 0.067]). More detailed information is provided below and summarized in \cref{tal0}.

\textbf{Dataset for Fundus Element Segmentation:}
For the segmentation task, 762 frames are randomly selected from the 208 videos and are pixel-level annotated for the eyeball and optic nerve sheath by two senior physicians using ITK-SNAP software \footnote{http://www.itksnap.org/}. To ensure non-overlapping patient data, these 762 annotated frames are divided into a training set and a test set at the 609:153 ratio, with all frames from each patient assigned to a single set exclusively.

\textbf{Dataset for Keyframe Identification:}
Addressing the technical challenges in keyframe identification described in \cref{Sec. 2.2}, two senior physicians perform textual annotations on 64 videos to identify classic keyframes and classic suboptimal frames. To ensure non-overlapping patient data, these annotated frames are partitioned into training and test sets: specifically, classic optimal and suboptimal frames from 40 videos are used to train a LDA algorithm for parameter determination, while frames from the remaining 24 videos serve exclusively for testing. The number of annotations per video varies from 1 to 10, depending on the video quality. This dataset is used to develop a robust keyframe identification algorithm.

\textbf{Dataset for ICP Grading:}
Based on patients’ contemporaneous clinical information, data from 64 patients’ 208 ocular fundus ultrasound videos are filtered to exclude those who have received mannitol or underwent abdominal shunting before lumbar puncture, as these interventions may distort intracranial pressure measurements. Ultimately, 102 paired binocular videos (51 left-eye and 51 right-eye) from 38 patients are selected for constructing the intracranial pressure grading model. Additionally, the independent test set derived from 18 patients undergoes the same screening criteria, resulting in 28 paired binocular videos (14 left-eye and 14 right-eye) from 13 patients. Notably, the clinical information requires lumbar puncture to obtain true ICP values, making such data difficult to acquire due to the invasive nature and clinical constraints of the procedure. This test set serves as an external validation cohort for evaluating the generalizability of the developed ICP grading model, and all data are acquired by a senior physician using different scanner models to ensure diversity.

\subsection{Evaluation metrics}
To quantitatively evaluate the performance of the fundus element segmentation module, this study employs the Dice Similarity Coefficient (Dice), calculated as follows:
\[
Dice(X,Y)=\frac{2|X \cap Y|}{|X|+|Y|},
\]
where $X$ and $Y$ denote the ground truth and predicted segmentation results, respectively.

Given the high similarity between adjacent frames in fundus ultrasound videos and the varying number of keyframes (ranging from 1 to 5 per video depending on quality), conventional classification metrics (e.g., accuracy, F1-score) are inadequate for evaluating the keyframe identification module. Therefore, we introduce the following metrics, mathematically defined using the indicator function \(I(\cdot)\) (where \(I(\cdot)\) returns 1 if the condition is true and 0 otherwise):  
\[
\text{Accuracy}_{\text{top1}} = \frac{1}{N} \sum_{i=1}^{N} I\left( \left| f_{\text{pred\_1}}^{(i)} - f_{\text{gt}}^{(i)} \right| \leq 2 \right),
\]  
\[
\text{Accuracy}_{\text{top3}} = \frac{1}{N} \sum_{i=1}^{N} I\left( \min_{k=1}^3 \left| f_{\text{pred\_k}}^{(i)} - f_{\text{gt}}^{(i)} \right| \leq 2 \right),
\]  
\[
\text{Accuracy}_{\text{top5}} = \frac{1}{N} \sum_{i=1}^{N} I\left( \min_{k=1}^5 \left| f_{\text{pred\_k}}^{(i)} - f_{\text{gt}}^{(i)} \right| \leq 2 \right),
\]  
where \(N\) is the total number of test samples, \(f_{\text{pred\_1}}^{(i)}\) is the frame index with the highest predicted score for the \(i\)-th sample, \(f_{\text{pred\_k}}^{(i)}\) denotes the indices of the top \(k\) frames with the highest predicted scores for the \(i\)-th sample (\(k = 1\) to \(5\) corresponds to the 1st to 5th ranked frame), and \(f_{\text{gt}}^{(i)}\) is the physician-annotated keyframe index.  

In the ICP grading module, this study employs four evaluation metrics to assess model performance. These metrics are mathematically defined as follows, using standard classification terminology:
\[
Accuracy=\frac{T P+T N}{T P+T N+F P+F N},
\]
\[
Precision=\frac{T P}{T P+F P},
\]
\[
Recall=\frac{T P}{T P+F N},
\]
\[
F1=\frac{2 T P}{2 T P+F P+F N}.
\]
where $TP$ denotes the number of true positive samples; $TN$, the number of true negative samples; $FP$, the number of false positive samples; and $FN$, the number of false negative samples.

\section{Results}
\begin{table*}[htbp]
\caption{Quantitative comparison of evaluation metrics for different classification models. Values are presented as mean±standard deviation. The best and second-best performances are highlighted in bold red and blue, respectively.} \label{tbl1}
\centering
\begin{tabular*}{\textwidth}{@{\extracolsep{\fill}}lcccc@{}}
\toprule
Model & Accuracy & Precision & Recall & F1-Score \\
\midrule
Logistic & $0.684\pm0.150$ & $0.729\pm0.123$ & $0.684\pm0.150$ & $0.680\pm0.151$ \\
Decision Tree & $0.665\pm0.105$ & $0.727\pm0.122$ & $0.665\pm0.105$ & $0.635\pm0.120$ \\
\textbf{Random Forest} & \bm{\textcolor{red}{$0.845\pm0.071$}} & \bm{\textcolor{red}{$0.838\pm0.104$}} & \bm{\textcolor{red}{$0.845\pm0.071$}} & \bm{\textcolor{red}{$0.831\pm0.086$}} \\
SVM & $0.665\pm0.138$ & $0.654\pm0.211$ & $0.665\pm0.138$ & $0.638\pm0.162$ \\
KNN & $0.724\pm0.101$ & $0.710\pm0.150$ & $0.724\pm0.101$ & $0.696\pm0.109$ \\
XGBoost & \textcolor{blue}{$0.805\pm0.055$} & \textcolor{blue}{$0.808\pm0.094$} & \textcolor{blue}{$0.805\pm0.055$} & \textcolor{blue}{$0.790\pm0.070$} \\
MLP & $0.685\pm0.100$ & $0.689\pm0.132$ & $0.685\pm0.100$ & $0.657\pm0.104$ \\
Bayes & $0.645\pm0.177$ & $0.697\pm0.213$ & $0.645\pm0.177$ & $0.649\pm0.187$ \\
\bottomrule
\end{tabular*}
\end{table*}

\subsection{Results of Fundus ultrasound videos processing stage}

In the fundus element segmentation module, the trained nn-UNet model achieves a Dice coefficient of 97.36\% for eyeball segmentation and 90.05\% for optic nerve sheath segmentation on 153 2D images in the test set, providing precise segmentation masks for subsequent modules. 

\begin{figure}[htbp]
\includegraphics[width=\columnwidth]{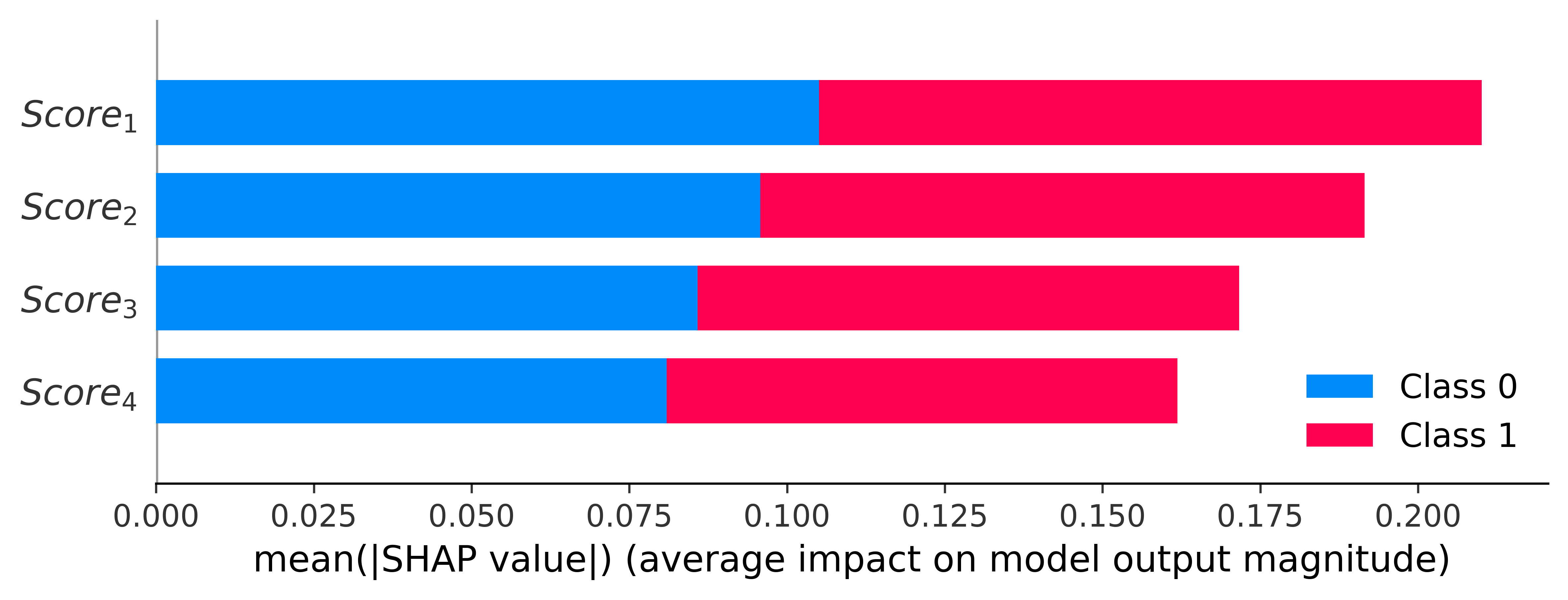}
    \caption{SHAP-based feature importance visualization of the four evaluation rules ($\text{score}_1$–$\text{score}_4$). The dual-color bars (blue: Class 0/suboptimal frames; red: Class 1/keyframes) quantify mean absolute SHAP values, reflecting the relative contribution of each rule to distinguishing keyframes.}\label{fig5}
\end{figure}

In the keyframe identification module, four evaluation rules are established, and their weight coefficients are determined via LDA based on text annotations of optimal and suboptimal frames from 40 video cases. Specifically, the weight coefficients are set as \(w_1 = -0.2873\), \(w_2 = 0.3585\), \(w_3 = 0.1115\), and \(w_4 = -0.1179\). Notably, the coefficients of $\operatorname{score}_2$ and $\operatorname{score}_3$ are negative, while those of $\operatorname{score}_1$ and $\operatorname{score}_4$ are positive, which aligns with our quantitative criteria. These coefficients reflect the relative importance of each rule in distinguishing keyframes, and their SHAP \citep{Lundberg_2017} importance rankings are shown in \cref{fig5}. After obtaining the coefficients, the algorithm-selected keyframes based on the final metric $\operatorname{score}_{\text{total}}$ are compared with physician-annotated keyframes in a 24-case test set. The results demonstrate an $Accuracy_{\text{top1}}$ of 0.417, an $Accuracy_{\text{top3}}$ of 0.958, and an $Accuracy_{\text{top5}}$ of 1.000.

\subsection{Results of ICP grading stage}

\begin{figure}[htbp]
\includegraphics[width=\columnwidth]{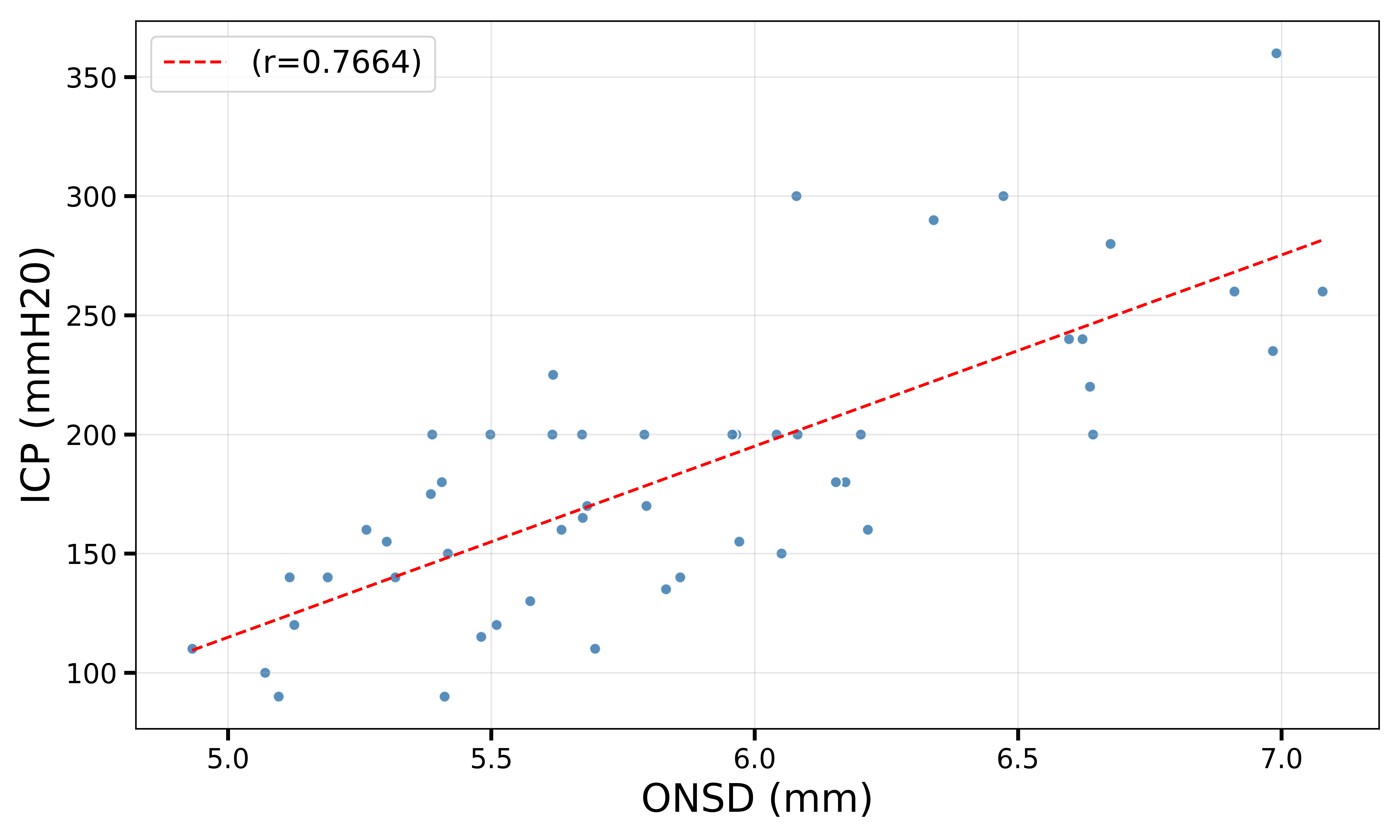}
    \caption{Scatter plot of the mean ONSD of both eyes and ICP from 102 paired binocular videos. The results show a significant positive correlation between ONSD and ICP (Pearson correlation coefficient r=0.7664, p<0.001).}\label{fig6}
\end{figure}

\begin{figure}[htbp]
\includegraphics[width=\columnwidth]{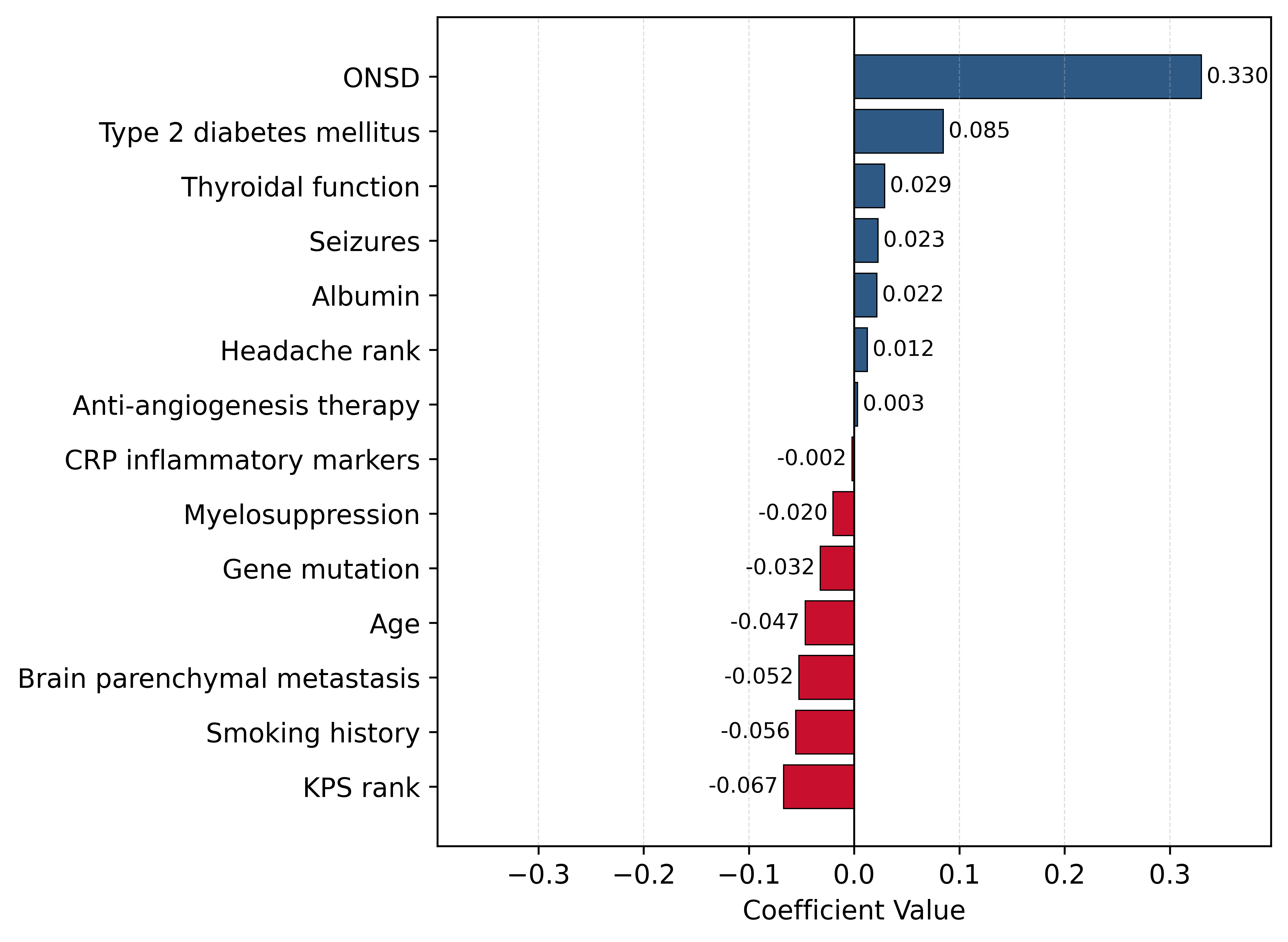}
    \caption{The 14 features selected by Lasso regression. The horizontal bar chart displays the regression coefficients associated with each feature, highlighting their importance in ICP grading prediction. Features with positive coefficients make positive contributions to the prediction, while those with negative coefficients exhibit negative correlations with ICP grading.}\label{fig7}
\end{figure}

In the ICP grading stage, we first perform correlation analysis between the mean ONSD of the left and right eyes from 102 paired binocular videos and patient ICP values, with the scatter plot shown in \cref{fig6}. The results indicate a significant Pearson correlation (r = 0.7664, \(p\)-value = \(5.5655 \times 10^{-11}\)), validating the feasibility of using ONSD for ICP grading. For clinical feature selection, Lasso regression is employed to screen 14 features from 50 candidates, with the contribution of each feature visualized in \cref{fig7}. Notably, ONSD exhibits the strongest correlation with patient ICP among all features. Finally, eight machine learning models—including Logistic regression, Decision Tree, Random Forest, SVM, KNN, XGBoost, MLP, and Bayesian classifier—are evaluated via five-fold cross-validation for ICP grading, with results summarized in \cref{tbl1}. The Random Forest classifier, demonstrating optimal performance, is selected as the ICP grading model for this study, and its five-fold cross-validation results are presented in \cref{tbl2}. Meanwhile, the five-fold cross-validation results and thresholds of ICP grading evaluation using threshold-based methods are summarized in \cref{tbl3}, wherein the optimal thresholds are determined by grid searching.

\begin{table}[htbp]
\caption{Model evaluation metric results of each fold in five-fold cross-validation for the Random Forest classifier.} \label{tbl2}
\centering
\begin{tabular*}{\columnwidth}{@{\extracolsep{\fill}}lcccc@{}}
\toprule
Fold & Accuracy & Precision & Recall & F1-Score \\
\midrule
0 & $0.727$ & $0.667$ & $0.727$ & $0.694$ \\
1 & $0.900$ & $0.917$ & $0.900$ & $0.897$ \\
\textbf{2} & \bm{$0.900$} & \bm{$0.920$} & \bm{$0.900$} & \bm{$0.900$} \\
3 & $0.900$ & $0.920$ & $0.900$ & $0.900$ \\
4 & $0.800$ & $0.767$ & $0.800$ & $0.764$ \\
\bottomrule
\end{tabular*}
\end{table}

\begin{table}[htbp]
\caption{Model evaluation metric results of each fold in five-fold cross-validation for conventional threshold-based method.}
\label{tbl3}
\centering
\begin{tabular*}{\columnwidth}{@{\extracolsep{\fill}}lcccccc@{}}
\toprule
Fold & Thresholds & Accuracy & Precision & Recall & F1 \\
\midrule
0 & $5.96$  $6.99$ & $0.636$ & $0.587$ & $0.636$ & $0.577$ \\
1 & $5.96$  $6.68$ & $0.600$ & $0.557$ & $0.600$ & $0.550$ \\
2 & $5.96$  $6.99$ & $0.600$ & $0.533$ & $0.600$ & $0.564$ \\
3 & $5.96$  $6.68$ & $0.700$ & \bm{$0.790$} & $0.700$ & $0.708$ \\
\textbf{4} & \bm{$5.96$}  \bm{$6.99$} & \bm{$0.800$} & $0.717 $& \bm{$0.800$} & \bm{$0.755$} \\
\bottomrule
\end{tabular*}
\end{table}

Notably, in the independent test set, the optimal Random Forest classifier achieves an accuracy of $0.786$, precision of $0.799$, recall of $0.786$, and F1-score of $0.788$; the threshold-based method (optimal thresholds $5.96$ and $6.99$) shows an accuracy of $0.429$, precision of $0.375$, recall of $0.429$, and F1-score of $0.397$. These findings strongly demonstrate the superior generalizability of our method.

\subsection{Ablation study}

\smallskip

\noindent\textbf{(1) The effect of Keyframe identification module}

An ablation study is conducted to verify the effectiveness of the proposed keyframe identification module. As shown in \cref{tbl4}, the first two rows present the performance results of directly using the narrowest frame (after excluding frames without optic nerve sheath visualization) and the widest frame of ultrasound videos as keyframes on the 24-case Keyframe Identification Module test dataset, respectively. The third row shows the results of our method. The findings indicate that although using the widest frame of the entire video as the keyframe achieves an $Accuracy_{\text{top1}}$ of 0.375 (consistent with medical knowledge), the proposed keyframe identification module significantly improves the three metrics by 6.17\%, 41.67\%, and 37.5\%, respectively.

\begin{table}[htbp]
\caption{Performance comparison of different frame selection methods on keyframe identification.} \label{tbl4}
\centering
\begin{tabular*}{\columnwidth}{@{\extracolsep{\fill}}p{0.25\columnwidth}ccc@{}}  
\toprule
Methods & $Accuracy_{\text{top1}}$& $Accuracy_{\text{top2}}$& $Accuracy_{\text{top3}}$\\
\midrule
Narrowest & $0.042$ & $0.042$ & $0.083$ \\
Widest  & $0.375$ & $0.542$ & $0.625$ \\
\textbf{Our method}    & \bm{$0.417$} & \bm{$0.958$} & \bm{$1.000$} \\
\bottomrule
\end{tabular*}
\end{table}

\smallskip

\noindent\textbf{(2) The effects of four rules of Keyframe identification}

We conduct ablation experiments to explore the effects of each selection rule in the keyframe identification module. As shown in \cref{tbl5}, the first four rows respectively present the experimental results of removing Rule 1, Rule 2, Rule 3, and Rule 4 individually. When all rules are activated synergistically (row 5), the three metrics, \(Accuracy_{\text{top1}}\), \(Accuracy_{\text{top3}}\), and \(Accuracy_{\text{top5}}\), all achieve the optimal performance. This demonstrates that the synergistic effect of the rules significantly enhances the efficiency of keyframe identification, verifying that each rule plays an irreplaceable role in the keyframe identification module and collectively provides theoretical and practical support for the accurate identification of keyframes.

\begin{table*}[htbp]

\caption{Quantitative results of ablation experiments on selection rules in keyframe identification module.}
\label{tbl5}
\centering
\begin{tabular*}{\textwidth}{@{\extracolsep{\fill}}ccccccc@{}}
\toprule
\textbf{Rule 1} & \textbf{Rule 2} & \textbf{Rule 3} & \textbf{Rule 4} & $Accuracy_{\text{top1}}$ & $Accuracy_{\text{top3}}$ & $Accuracy_{\text{top5}}$ \\
\midrule
\ding{53} & \ding{51} & \ding{51} & \ding{51} & $0.417$ & $0.916$ & $1.000$ \\
\ding{51} & \ding{53} & \ding{51} & \ding{51} & $0.417$ & $0.75$ & $0.833$ \\
\ding{51} & \ding{51} & \ding{53} & \ding{51} & $0.375$ & $0.708$ & $0.833$ \\
\ding{51} & \ding{51} & \ding{51} & \ding{53} & $0.417$ & $0.916$ & $1.000$ \\
\ding{51} & \ding{51} & \ding{51} & \ding{51} & \bm{$0.417$} & \bm{$0.958$} & \bm{$1.000$} \\  
\bottomrule
\end{tabular*}
\end{table*}

\smallskip

\noindent\textbf{(3)The effects of components in ONSD measurement module}

We also conduct ablation experiments on the strategies within the ONSD measurement module to assess their influences on the correlation with ICP. As presented in \cref{tab6}, two scenarios are examined: without the IQR outlier - removal method (the first three rows), where the narrowest, average, and widest ONSD values from keyframe set are taken as the final video - level ONSD; and with the IQR outlier - removal method (the last three rows), where the same process is repeated after excluding extreme values via the IQR. Notably, the “Max (with IQR)” strategy yields the highest Pearson correlation (\(r = 0.7664\), \(p = 5.5655\times10^{-11}\)). This means that integrating IQR - based outlier removal with the measurement of the widest ONSD as the final video - level ONSD has the closest relationship with ICP.

\begin{table}[htbp]
\caption{Ablation experiments on ONSD measurement strategies. IQR =  Interquartile Range; $r$ = Pearson correlation coefficient; $p$-value = statistical significance.}
\label{tab6}
\centering
\begin{tabular*}{\columnwidth}{@{\extracolsep{\fill}}lcc@{}}
\toprule
\textbf{Strategy} & \textbf{Pearson $r$} & \textbf{$p$-value} \\
\midrule
Min (w/o IQR)         & $0.5671$ & $1.4339 \times 10^{-5}$ \\
Ave (w/o IQR)         & $0.7198$ & $2.6478 \times 10^{-9}$ \\
Max (w/o IQR)         & $0.7290$ & $1.3151 \times 10^{-9}$ \\
\midrule
Min (with IQR)        & $0.6701$ & $7.5044 \times 10^{-8}$ \\
Ave (with IQR)        & $0.7432$ & $4.2203 \times 10^{-10}$ \\
\textbf{Max (with IQR)}        & \bm{$0.7664$} & \bm{$5.5655 \times 10^{-11}$} \\
\bottomrule
\end{tabular*}
\end{table}

\smallskip

\noindent\textbf{(4)The effects of ONSD and clinical features on ICP grading}

To evaluate the individual and combined effects of ONSD and clinical features on ICP grading, ablation experiments are conducted on the independent test set. The results in \cref{tbl7} show that ONSD alone achieves an accuracy of $0.429$, clinical features alone reach $0.714$, and the combination of both yields the highest accuracy of $0.786$. This indicates the superiority of integrating ONSD and clinical features in ICP grading prediction.

\begin{table}[htbp]
\caption{Ablation experiments for evaluating ONSD and clinical features in ICP grading. CF = Clinical Features.}
\label{tbl7}
\centering
\begin{tabular*}{\columnwidth}{@{\extracolsep{\fill}}cccccc@{}}
\toprule
ONSD & CF & Accuracy & Precision & Recall & F1 \\
\midrule
\ding{51} & \ding{53} & $0.429$ & $0.375$ & $0.429$ & $0.397$ \\
\ding{53} & \ding{51} & $0.714$ & $0.693$ & $0.714$ & $0.695$ \\
\ding{51} & \ding{51} & \bm{$0.786$} & \bm{$0.799$} & \bm{$0.786$} & \bm{$0.788$} \\
\bottomrule
\end{tabular*}
\end{table}


\section{Conclusion and discussion}

In this paper, we develop a fully automatic framework for identifying keyframes and measuring ONSD in ultrasound videos, and predicting ICP grades by integrating clinical information and ONSD. The framework consists of two stages: fundus ultrasound video processing and ICP grading. Under this framework, clinicians can automatically obtain ONSD values and ICP grades by providing fundus ultrasound videos and corresponding clinical data, enabling timely screening of patients with high ICP at risk of brain herniation. Our ICP grading model achieves an accuracy of $0.845 \pm 0.071$ on the validation set and $0.786$ on an independent test
set, demonstrating significant superiority over conventional threshold-based method ($0.637 \pm 0.111$ validation accuracy, $0.429$ test accuracy). Experimental results show that the proposed framework has great potential to reduce the need for invasive lumbar puncture. To our knowledge, this is the first work to identify keyframes in ocular
fundus ultrasound videos in strict accordance with the international consensus statement on optic nerve sheath diameter imaging and measurement.

There are several possible limitations in our study. First, the keyframe scoring model (\cref{Eq.1}) assumes a linear relationship between the four indicator scores and the overall keyframe score, which may not adequately capture the nonlinear dependencies inherent in their real-world associations. Future research could explore nonlinear models to better characterize the complex mapping from indicator scores to the composite keyframe score. Second, when training the ICP grading model, we exclude patients who had taken mannitol or undergone abdominal shunting before lumbar puncture, reducing the sample size from 64 patients to 38 patients. The limited sample size may affect the generalization ability of our ICP grading model. As the number of patients accumulates, we will further optimize this grading model in future work.

\printcredits

\section*{Declaration of competing interest}
The authors declare that they have no known competing financial interests or personal relationships that could have appeared to influence the work reported in this paper.

\section*{Acknowledgments}

This work is supported by National Natural Science Foundation of China (Nos. 12301661,12431018), Nanjing Drum Tower Hospital Clinical Research Special Fund Project (No. 2024-LCYJ-PY-60) and Nanjing Drum Tower Hospital New Technology Development Project (No. XJSFZLX202423). Corresponding Ethics Number in Clinical Trials: NCT07005791.

\section*{Data and code availability}
Our data and code will be publicly aviliable upon acceptance.










\bibliographystyle{cas-model2-names}

\bibliography{cas-refs}



\end{document}